\documentclass[11pt]{article}

\usepackage[final]{acl}

\usepackage{times}
\usepackage{amsmath}
\usepackage{amssymb}
\usepackage{latexsym}
\usepackage{makecell}
\usepackage{booktabs}
\usepackage{multirow}
\usepackage{graphicx}
\usepackage{xcolor}
\usepackage{booktabs}
\usepackage{colortbl}
\usepackage[table]{xcolor}
\usepackage[utf8]{inputenc}    
\usepackage{tabularx}
\usepackage[table]{xcolor}
\usepackage{booktabs}
\usepackage{multirow} 
\usepackage{makecell} 
\usepackage{algorithm}
\usepackage{algorithmic}
\usepackage{booktabs}   
\usepackage{tabularx}   
\usepackage{adjustbox}  
\usepackage{amsmath}    
\usepackage{soul}

\PassOptionsToPackage{table}{xcolor}
\definecolor{green1}{RGB}{193, 229, 245}
\definecolor{red1}{RGB}{242, 207, 238}
\definecolor{orange1}{RGB}{246, 194, 66}
\definecolor{darkgreen}{rgb}{0.0, 0.5, 0.0}
\definecolor{darkred}{rgb}{0.7, 0.0, 0.0}

\usepackage{pifont}
\definecolor{DarkGreen}{RGB}{9, 136, 66} %
\definecolor{Maroon}{RGB}{238, 49, 49}    %
\newcommand{\greenCheckmarkBold}{\textcolor{DarkGreen}{\ding{51}}} %
\newcommand{\redXSolidBrush}{\textcolor{Maroon}{\ding{55}}}    %

\usepackage[T1]{fontenc}

\usepackage[utf8]{inputenc}

\usepackage{microtype}

\usepackage{inconsolata}

\usepackage{graphicx}
\definecolor{lightblue}{RGB}{220,230,255}
\title{Prefix-Adaptive Block Diffusion for Efficient Document Recognition}

\author{
\textbf{Mingxu Chai}\textsuperscript{1,2,3}\thanks{Equal contribution.}, 
\textbf{Ziyu Shen}\textsuperscript{1}\footnotemark[1], 
\textbf{Chenyu Liu}\textsuperscript{1}, \\
\textbf{Jihua Kang}\textsuperscript{4},  
\textbf{Tao Gui}\textsuperscript{1,2}, 
\textbf{Qi Zhang}\textsuperscript{1,3,5}\thanks{Corresponding author.} \\
\textsuperscript{1}College of Computer Science and Artificial Intelligence, Fudan University \\
\textsuperscript{2}Shanghai Innovation Institute
\textsuperscript{3}Shanghai Artificial Intelligence Laboratory \\
\textsuperscript{4}ByteDance, LarkAI
\textsuperscript{5}Shanghai Key Lab of Intelligent Information Processing \\
\texttt{qz@fudan.edu.cn}
}

\begin{document}
\maketitle
\begingroup
\renewcommand\thefootnote{}
\footnotetext{Code: \url{https://github.com/SII-sc22mc/PA-BDM}.}
\addtocounter{footnote}{-1}
\endgroup

\begin{abstract}
Block Diffusion Models (BDMs) support parallel generation, flexible-length output, and KV caching, making them promising for efficient document parsing.
However, existing BDMs bind denoising and cache commitment to fixed block boundaries: parallelism shrinks during intra-block denoising, while generated tokens cannot be cached until the whole block is completed.
Moreover, intra-block bidirectional denoising conflicts with inter-block autoregression, creating inconsistent information flow that can challenge structure-sensitive recognition.
We propose the Prefix-Adaptive Block Diffusion Model (PA-BDM), which replaces intra-block bidirectional denoising with causal denoising from prefix to suffix and treats the block size as a maximum candidate range rather than a fixed commitment unit.
PA-BDM uses Confidence-gated Structural Loss (CSL) to build low-entropy prefixes before extending training to longer continuations.
During inference, Progressive Prefix Commitment (PPC) then dynamically commits the longest reliable prefix into the KV cache and resets the next candidate range from the updated prefix, restoring a large parallel decoding space at each step.
Experiments show that the 3B PA-BDM achieves higher recognition scores on several benchmarks and improves inference throughput by 71.6\% over the 2.5B MinerU-Diffusion.
\end{abstract}

\section{Introduction}

\begin{figure*}
  \centering
  \includegraphics[width=1.0\linewidth, trim=0 320 0 0, clip]{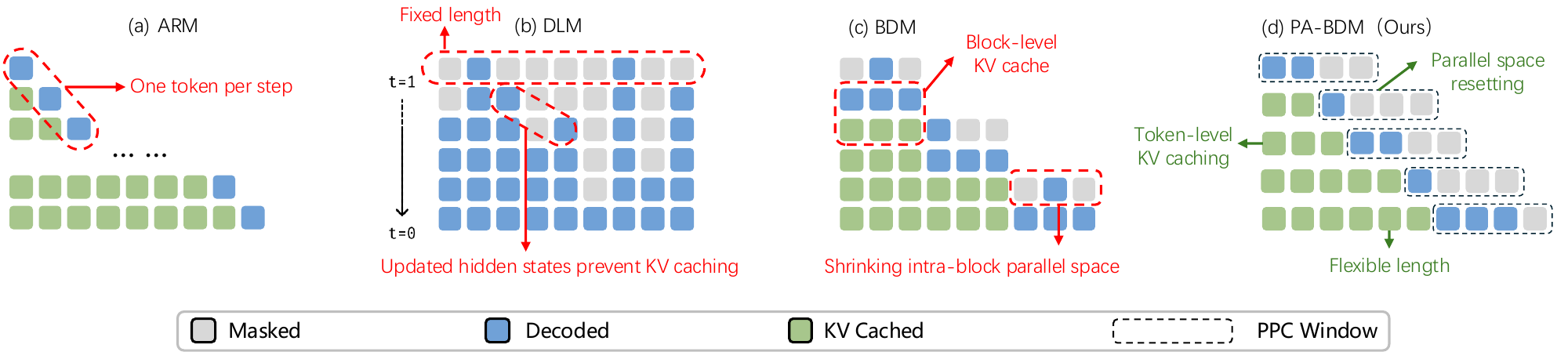}
  \caption{
Unlike standard block diffusion models that cache only after completing an entire block, our method treats each block as a candidate generation range and progressively commits reliable prefixes into the KV cache, enabling adaptive generation and caching granularity.
\textit{Note that standard diffusion models do not naturally support exact KV caching, while recent methods modify training or inference to approximate cache-like behavior.}
}
  \label{fig:intro1}
\end{figure*}

Document parsing aims to recognize document images into machine-readable formats~\cite{zhang2025documentparsingunveiledtechniques}. 
Mainstream methods are based on Autoregressive Models (ARMs), which have achieved substantial progress~\cite{niu2025mineru25decoupledvisionlanguagemodel,cui2025paddleocrvlboostingmultilingualdocument}, yet their strictly token-by-token generation paradigm limits inference efficiency.
To address this, recent studies have explored various parallel generation paradigms~\cite{du2025mdiff4strmaskdiffusionmodel,duan2026glmocrtechnicalreport}.
Among them, Block Diffusion Models (BDMs) offer a promising direction by generating blocks autoregressively while denoising tokens inside each block in parallel, thereby improving decoding parallelism while retaining flexible-length generation and KV-cache reuse~\cite{man2026dododiscreteocrdiffusion}.

However, when applied to document recognition, the standard BDM formulation reveals limitations in both efficiency and structural modeling.
First, standard BDMs rely on predefined block boundaries and use them as both local denoising ranges and cache-commitment units, as shown in Fig.~\ref{fig:intro1}(c). As a result, generated tokens can be written into the KV cache only after the whole block is completed, making it difficult to promptly reuse reliable intermediate predictions. Meanwhile, as intra-block denoising proceeds, the number of remaining masked tokens gradually decreases, reducing the effective parallel decoding space.
Second, standard BDMs introduce inconsistent information flow between intra-block and inter-block modeling: tokens within the same block can condition on each other bidirectionally, while cross-block generation still follows a left-to-right autoregressive order. For tasks mainly driven by global semantics, such bidirectional context may be beneficial, since output quality does not always depend on the exact structural position of each token. However, document recognition requires precise reconstruction of discrete token sequences from visual content, and structured outputs such as LaTeX and HTML are especially sensitive to token order and structural boundaries. Therefore, when similar local structures receive different conditioning patterns due to their positions relative to block boundaries, the model may find it harder to learn consistent structural generation patterns.

Based on these observations, we propose the Prefix-Adaptive Block Diffusion Model (PA-BDM).
PA-BDM replaces intra-block bidirectional denoising with causal denoising inside each candidate block, aligning intra-block information flow with inter-block autoregressive progression.
This reduces boundary-dependent conditioning and makes reliable candidate prefixes valid for KV-cache reuse.
Thus, the block size is no longer an indivisible generation and commitment unit, but serves as the maximum candidate range of each forward pass.
During inference, Progressive Prefix Commitment (PPC) dynamically commits the longest contiguous reliable prefix from each causally constrained parallel prediction and resets the next candidate range from the updated prefix.
This enables timely reuse of reliable predictions and restores a large parallel decoding space at each step, as shown in Fig.~\ref{fig:intro1}(d).
During training, PA-BDM uses Confidence-gated Structural Loss (CSL) to match causal candidate-block denoising.
CSL adjusts supervision according to prefix confidence, encouraging reliable structural prefixes before longer continuations and reducing noisy supervision from unstable prefix states.

We instantiate PA-BDM on DiffusionVL~\cite{zeng2026diffusionvltranslatingautoregressivemodels} at multiple model scales and evaluate it on text~\cite{ouyang2025omnidocbenchbenchmarkingdiversepdf}, table~\cite{zheng2020globaltableextractorgte}, formula~\cite{wang2024unimernetuniversalnetworkrealworld}, and diagram recognition tasks~\cite{pan2024flowlearnevaluatinglargevisionlanguage}.
Experimental results show that PA-BDM improves recognition accuracy over BDM baselines, with particularly strong gains on complex formula recognition, while PPC substantially improves inference efficiency through adaptive prefix commitment and timely KV-cache reuse.
Overall, PA-BDM achieves a stronger speed--accuracy trade-off than diffusion-based baselines on several structure-sensitive benchmarks, while delivering a 71.6\% speedup over MinerU-Diffusion and around 8$\times$ higher throughput than the compared ARM recognizers.
The main contributions are:

\begin{itemize}
    \item We identify fixed block-level commitment and inconsistent intra-/inter-block information flow as key limitations of BDMs for structure-sensitive document recognition.

    \item We propose PA-BDM, a prefix-adaptive BDM framework that redefines the block as a maximum candidate generation range rather than a fixed generation and cache-commitment unit.

    \item We introduce PPC to dynamically commit reliable prefixes, enable timely KV-cache reuse, and reset the candidate range to recover parallel decoding space.

    \item Experiments show that PA-BDM improves both accuracy and inference throughput over comparable DLM and BDM baselines.
\end{itemize}

\section{Related Work}
\label{sec:related_work}

\subsection{Diffusion Models for Efficient Decoding}
\label{sec:related_work_dlm}

Diffusion Language Models (DLMs) enable parallel token prediction and provide an alternative to autoregressive decoding~\cite{nie2025largelanguagediffusionmodels, ye2025dream7bdiffusionlarge}. 
However, standard DLMs usually denoise within a fixed-length token space, which limits flexible-length generation, and repeatedly update decoded hidden states, making exact KV caching difficult.
Recent methods improve DLM inference through approximate caching~\cite{wu2025fastdllmtrainingfreeaccelerationdiffusion}, prefix KV mechanisms~\cite{li2025lavidalargediffusionlanguage}, but they mainly accelerate fixed-space denoising rather than progressively converting reliable predictions into reusable causal prefixes.
Block Diffusion~\cite{arriola2025blockdiffusioninterpolatingautoregressive} combines inter-block autoregression with intra-block parallel denoising, supporting flexible-length generation and block-level KV caching.
Nevertheless, its generation and cache commitment are still tied to whole-block completion.
PA-BDM removes this fixed block-level granularity by treating the block size as a maximum candidate range and adaptively committing reliable prefixes for KV reuse.
Unlike speculative decoding~\cite{leviathan2023fastinferencetransformersspeculative}, which relies on draft-and-verify prediction with an auxiliary draft model, PPC performs confidence-based prefix commitment inside a single model.

\subsection{Document Parsing Models}
\label{sec:related_work_doc_parsing}
Autoregressive vision-language models have become a dominant paradigm for document parsing~\cite{blecher2023nougatneuralopticalunderstanding, wei2024generalocrtheoryocr20, feng2025dolphindocumentimageparsing, niu2025mineru25decoupledvisionlanguagemodel, cui2025paddleocrvlboostingmultilingualdocument}. 
Their causal token-by-token generation provides stable prefix conditioning, which is important for structure-sensitive outputs such as formulas and tables, but also limits inference efficiency.
Recent work has therefore explored more efficient generation paradigms.
GLM-OCR improves efficiency by predicting multiple tokens in parallel under a globally causal attention pattern~\cite{duan2026glmocrtechnicalreport}, indicating that parallel decoding does not necessarily require bidirectional attention.
Diffusion-based methods such as MDiff4STR and MinerU-Diffusion pursue higher parallelism through iterative denoising~\cite{du2025mdiff4strmaskdiffusionmodel,dong2026minerudiffusionrethinkingdocumentocr}.
DODO further introduces block diffusion into document recognition~\cite{man2026dododiscreteocrdiffusion}. 
Despite this progress, existing diffusion-based methods are either mainly evaluated on text-centric recognition or still show accuracy gaps against strong autoregressive baselines on complex structured outputs.

\section{Method}
\label{sec:method}

We first review the standard vision-language block diffusion model formulation used in DiffusionVL~\cite{zeng2026diffusionvltranslatingautoregressivemodels} in Sec.~\ref{sec:preliminaries}.
We then introduce the overall PA-BDM framework in Sec.~\ref{sec:pabdm}, followed by Progressive Prefix Commitment (PPC) in Sec.~\ref{sec:ppc}.

\subsection{Preliminaries}
\label{sec:preliminaries}

Given an input image $I$ and a text prompt $T$, DiffusionVL generates a variable-length response sequence
$x=\{x_1,\ldots,x_N\}$.
For block-wise modeling, we consider a block-aligned length
$L=\lceil N/D\rceil D$, where $D$ is the block size.
The response positions are organized into $B=L/D$ blocks, with the last block containing auxiliary positions if $N$ is not divisible by $D$:
\[
X_b=\{x_{(b-1)D+1},\ldots,x_{bD}\}, \quad b=1,\ldots,B.
\]
Only the valid response positions are used for supervision and evaluation.

During training, DiffusionVL applies block-wise noise.
For each block $X_b$, a noise level $t_b\sim U(0,1)$ is sampled, and each token in the block is independently replaced by $[\mathrm{MASK}]$ with probability $t_b$:
\begin{equation}
\tilde{x}_{i} =
\begin{cases}
[\mathrm{MASK}], & \text{with probability } t_{\beta(i)}, \\
x_i, & \text{with probability } 1-t_{\beta(i)},
\end{cases}
\end{equation}
where $\beta(i)=\lceil i/D\rceil$ denotes the block index of position $i$.

The attention pattern is semi-autoregressive.
For response-token positions $i$ and $j$, the standard BDM attention mask is
\begin{equation}
A^{\mathrm{BDM}}_{ij} =
\begin{cases}
1, & \beta(j) < \beta(i), \\
1, & \beta(j) = \beta(i), \\
0, & \text{otherwise}.
\end{cases}
\end{equation}
Thus, tokens in previous blocks are visible, tokens within the same block are denoised bidirectionally, and future blocks are masked.

During inference, visual and prompt tokens are first encoded into an initial cache $C_0$.
For each block $b$, the model appends $D$ mask tokens and iteratively denoises them conditioned on the previous cache $C_{b-1}$.
Only after the whole block is completed are its predicted tokens $\hat{X}_b$ committed and their KV states materialized:
\begin{equation}
C_b = \mathrm{Append}(C_{b-1}, \hat{X}_b),
\end{equation}
where $\mathrm{Append}$ denotes appending the KV states of the completed block.
Therefore, standard BDM-style decoding supports block-level KV-cache reuse, but its generation and cache-commitment granularity are tied to whole-block completion.
Moreover, as denoising proceeds within the same block, fewer $[\mathrm{MASK}]$ tokens remain to be updated, so the effective parallel decoding space gradually shrinks before the block can be committed.

\begin{figure*}[ht]
  \centering
  \includegraphics[width=1.0\linewidth, trim=0 275 25    0, clip]{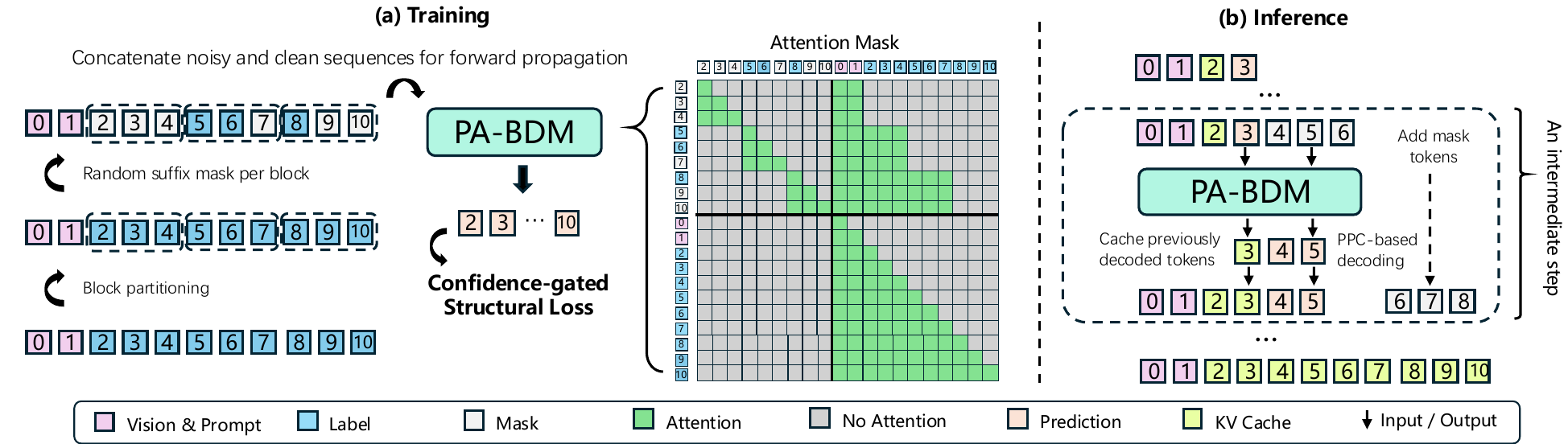}
  \caption{
\textbf{Training and inference of PA-BDM.}
(a) During training, PA-BDM concatenates noisy and clean sequences, applies causal block attention, and uses CSL to supervise as many masked tokens as allowed by prefix confidence.
(b) During inference, PA-BDM treats the block size as a maximum candidate range. PPC selects a committed prefix, materializes its KV states while predicting the next candidate range, and resets unresolved positions with new mask tokens, enabling adaptive generation and caching granularity.
}
  \label{fig:intro2}
\end{figure*}

\subsection{Prefix-Adaptive Block Diffusion Model}
\label{sec:pabdm}

Instead of treating the block size $D$ as an indivisible commitment unit, PA-BDM uses it as the maximum candidate generation range and commits only a reliable prefix at each generation round.

During training, we follow DiffusionVL and build a concatenated input $\xi=[\tilde{x},x]$ from noisy and clean response sequences, as shown in Fig.~\ref{fig:intro2}(a).
The clean branch is used to construct block-wise training context for the noisy branch under controlled attention.
PA-BDM replaces intra-block bidirectional denoising with causal denoising:
\begin{equation}
A^{\mathrm{PA}}_{ij} =
\begin{cases}
1, & \beta(j) < \beta(i), \\
1, & \beta(j) = \beta(i) \text{ and } j \le i, \\
0, & \text{otherwise}.
\end{cases}
\end{equation}
This mask allows each position to attend to previous blocks and its left-side context within the current block.
Therefore, response-token attention follows a prefix-to-suffix order across both intra-block and inter-block positions, so a reliable prefix inside a candidate block can be treated as a valid continuation for KV-cache reuse.

Although PA-BDM uses causal attention, its training is not equivalent to standard autoregressive training.
The left context of a masked position may still contain uncertain masked tokens, so directly supervising the whole masked suffix can introduce noisy gradients from continuations conditioned on unstable prefix states.
To address this, PA-BDM uses Confidence-gated Structural Loss (CSL) to make supervision prefix-aware.

For notation, let $x_{b,k}=x_{(b-1)D+k}$ denote the clean target token at the $k$-th position of block $b$.
For each block $X_b$, we sample a suffix start $u_b$ and define the masked suffix as
$
\Omega_b=\{u_b,\ldots,D\}.
$
For each masked position $k\in\Omega_b$, we compute the gold-token confidence
$
q_{b,k}=P_\theta(x_{b,k}\mid \xi),
$
where $q_{b,k}$ is detached from gradient computation.
Let $h_b$ be the first position in $\Omega_b$ whose confidence is below the threshold $\tau$:
\begin{equation}
h_b=\min\{k\in\Omega_b \mid q_{b,k}<\tau\}.
\end{equation}
If no such position exists, all positions in $\Omega_b$ are supervised. Otherwise, CSL supervises positions only up to $h_b$:
\begin{equation}
S_b =
\begin{cases}
\Omega_b, & \text{if no } h_b \text{ exists},\\
\{u_b,\ldots,h_b\}, & \text{otherwise}.
\end{cases}
\end{equation}
Only positions in $\{S_b\}_{b=1}^{B}$ contribute to the cross-entropy loss, while the remaining masked positions are excluded from gradient computation.

CSL is not an easy-token filtering strategy.
When an early continuation token has low confidence, it is still included in $S_b$ and becomes the current learning frontier.
Therefore, a degenerate behavior that only predicts the first masked token confidently is not optimal, because the next low-confidence token will repeatedly receive direct cross-entropy supervision until its confidence improves.
The randomly sampled suffix start $u_b$ further allows different relative positions to appear near the beginning of the masked suffix and receive direct supervision.
Since the confidence scores are detached, CSL provides no differentiable incentive to lower confidence in order to shorten the supervised range.

During inference, at generation round $r$, PA-BDM appends $D$ temporary mask tokens after the current prefix and predicts
$
\hat{X}^{(r)}=\{\hat{x}^{(r)}_1,\ldots,\hat{x}^{(r)}_D\}.
$
It commits only a prefix $\hat{X}^{(r)}_{1:\ell_r}$ with $0<\ell_r\le D$ and discards the unresolved suffix.
Therefore, $D$ defines the maximum candidate range, while $\ell_r$ determines the actual generation and caching granularity.
Sec.~\ref{sec:ppc} describes how $\ell_r$ is selected and cached.

\subsection{Progressive Prefix Commitment}
\label{sec:ppc}

Progressive Prefix Commitment (PPC) determines the committed length $\ell_r$ at each generation round.
Given the current candidate prediction
$
\hat{X}^{(r)}=\{\hat{x}^{(r)}_1,\ldots,\hat{x}^{(r)}_D\},
$
we compute the confidence of each predicted token from the current forward pass:
\begin{equation}
c_k^{(r)}
=
P_\theta\left(\hat{x}^{(r)}_k \mid \mathcal{H}^{(r)}\right),
\quad k=1,\ldots,D,
\end{equation}
where $\mathcal{H}^{(r)}$ denotes the actual context used to predict the $r$-th candidate range.

PPC scans the candidate range from left to right and commits the longest high-confidence prefix.
Once a low-confidence token is encountered, positions to its right remain unresolved even if they have high individual confidence.
To ensure monotonic progress, PPC commits one token when the first token is already below the threshold.
Let $m_r$ be the first low-confidence position, if it exists:
\begin{equation}
m_r = \min \{ k\in\{1,\ldots,D\} \mid c_k^{(r)} < \tau \}.
\end{equation}
The committed length is
\begin{equation}
\ell_r =
\begin{cases}
D, & \text{if no such } m_r \text{ exists}, \\
\max(1, m_r-1), & \text{otherwise},
\end{cases}
\end{equation}
where $\tau$ is the confidence threshold.
The selected prefix is
$
\bar{X}^{(r)}=\hat{X}^{(r)}_{1:\ell_r}.
$

This left-contiguous commitment is valid because PA-BDM uses causal attention inside the candidate range.
Tokens in $\bar{X}^{(r)}$ depend only on the previous cache and earlier tokens in the same prefix, not on the unresolved suffix.
Therefore, once $\bar{X}^{(r)}$ is fed back as clean input, its KV states can be safely reused as prefix context.

As shown in Fig.~\ref{fig:intro2}(b), the current denoising pass only decides which token identities to commit, since it predicts from mask inputs.
PA-BDM materializes the selected prefix in the next forward pass together with the next candidate prediction, avoiding an extra cache-only pass.
Let $C^{(0)}=C_0$ be the initial cache of visual and prompt tokens.
After $\bar{X}^{(r)}$ is selected, the next forward pass takes the previous cache $C^{(r-1)}$, the selected prefix $\bar{X}^{(r)}$, and a new temporary candidate range $\tilde{X}^{(r+1)}$:
\begin{equation}
\left(C^{(r)}, \hat{X}^{(r+1)}\right)
=
\mathcal{F}_{\theta}\big(C^{(r-1)}, \bar{X}^{(r)}, \tilde{X}^{(r+1)}\big).
\end{equation}
Here, $C^{(r)}$ appends only the materialized KV states of $\bar{X}^{(r)}$, while $\hat{X}^{(r+1)}$ is the prediction over the new candidate range and is not cached.
Thus, PA-BDM combines prefix materialization and candidate-range prediction in one forward pass, allowing decoding to reset a full candidate range after each committed prefix rather than spending later steps on a shrinking residual suffix.

\section{Experiments}

\begin{table*}[t!]
    \centering
    \begin{adjustbox}{max width=\textwidth}
    \begin{tabular}{l|cccc|cccc}
      \toprule
         & PA-BDM(ours) & DiffusionVL & MinerU-Diffusion & LaViDa & MonkeyOCR-Pro & MinerU2.5-Pro & Dolphinv2 & Qwen2.5VL  \\
      \midrule
      Type & BDM & BDM & DLM & DLM & ARM & ARM & ARM & ARM \\
      Size & 3B & 3B & 2.5B & 3B & 3B & 1.2B & 3B & 72B \\
      TPS & 267.2 & 92.3 &155.7  &72.5  & 32.3 & 48.4 & 34.1 & - \\
      Mem & 7.3 & 7.2 &5.8  &8.0  & 7.2 & 2.3 & 7.3 & - \\
      \midrule
      \multicolumn{9}{c}{\textit{Formula ($\text{CDM} \uparrow$)}}\\
      \midrule
      SPE & \textbf{98.7} & 95.3 & 96.8 & 91.2 & 97.6 & \underline{99.4} & 98.1 & 96.2\\
      SCE & \textbf{94.3} & 91.2 & 92.0 & 83.9 & 94.9 & \underline{97.0} & 95.5 & 95.5\\
      CPE & \textbf{94.7} & 64.3 & 91.6 & 43.7 & 91.4 & \underline{98.9} & 88.1 & 88.9\\
      HWE & \textbf{93.8} & 86.7 & 91.6 & 84.6 & 92.2 & \underline{95.3} & 90.5 & 91.8\\
      \midrule
      \multicolumn{9}{c}{\textit{Text ($\text{Edit} \downarrow$)}}\\
      \midrule
      DocLaynet & \textbf{0.087} & 0.135 & 0.112 & 0.105 & \underline{0.080} & 0.084 & 0.102 & 0.096\\
      OmniDoc & 0.093 & 0.121 & \textbf{0.085} & 0.093 & 0.071 & \underline{0.064} & 0.078 & 0.073\\
      \midrule
      \multicolumn{9}{c}{\textit{Diagram ($\text{F1} \uparrow$)}}\\
      \midrule
      FlowLearn & \textbf{\underline{90.4}} & 63.7 & - & 77.3 & - & - & - & 54.8\\
      \midrule
      \multicolumn{9}{c}{\textit{Table ($\text{TEDS} \uparrow$)}} \\
      \midrule
      PubtableNet & \textbf{89.6} & 81.4 & 84.2 & 69.4 & 87.4 & 90.1 & \underline{90.6} & 84.3\\
      FinTabNet & \textbf{88.3} & 83.1 & 86.7 & 71.1 & 86.4 & \underline{95.1} & 87.1 & 82.9\\
      \bottomrule
    \end{tabular}
    \end{adjustbox}
    \caption{
Main comparison on document recognition benchmarks.
\textbf{Bold} indicates the best diffusion-based result, and \underline{underline} indicates the best overall result.
TPS denotes generated tokens per second averaged over all non-diagram benchmarks, and Mem denotes peak GPU memory in GB.
BDM, DLM, and ARM refer to block diffusion, diffusion language, and autoregressive models, respectively.
MinerU-Diffusion and LaViDa are DLM-based models with block-wise inference decoding.
}
    \label{tab:recog}
\end{table*}

\textbf{Implementation Details.}
\label{sec:implementation_details}
Unless otherwise specified, we set the maximum candidate block size to 32 and use a confidence threshold of 0.95 for both CSL and PPC-based prefix caching.
All speed measurements are conducted on a single NVIDIA RTX 4090 GPU with batch size 1.
Detailed training configurations are provided in Appendix~\ref{appendix:training_configuration}.

\noindent
\textbf{Evaluation Benchmarks and Metrics.}
We evaluate PA-BDM on recognition benchmarks covering four task types:
(1) text, including OCR-blocks from DocLayNet~\cite{Pfitzmann_2022} and OmniDoc~\cite{ouyang2025omnidocbenchbenchmarkingdiversepdf};
(2) formula, using UniMER-Test~\cite{wang2024unimernetuniversalnetworkrealworld} with Simple Printed Expressions (SPE), Complex Printed Expressions (CPE), Screen-Captured Expressions (SCE), and Handwritten (HWE);
(3) table, including PubTabNet~\cite{zhong2020imagebasedtablerecognitiondata} and FinTabNet~\cite{zheng2020globaltableextractorgte};
and (4) diagram, using FlowLearn~\cite{pan2024flowlearnevaluatinglargevisionlanguage}.
We report Edit Distance (Edit) for text, Character Detection Matching (CDM)~\cite{wang2024cdmreliablemetricfair} for formulas, Tree-Edit-Distance-based Similarity (TEDS)~\cite{zhong2020imagebasedtablerecognitiondata} for tables, and F1 for diagrams.
We also use ACC as an aggregate evaluation metric in later analyses, with its computation and detailed descriptions of all metrics provided in Appendix~\ref{appendix:metrics}.

\noindent
\textbf{Baselines.}
We compare PA-BDM with both diffusion and ARM baselines.
For controlled comparisons, we re-train LaVida~\cite{li2025lavidalargediffusionlanguage} and DiffusionVL~\cite{zeng2026diffusionvltranslatingautoregressivemodels} on the same training data as PA-BDM, with matched visual and language model sizes.
In these controlled comparisons, LaVida serves as the DLM baseline, while DiffusionVL serves as the BDM baseline.
For comparison with existing public systems, we directly evaluate released models without additional fine-tuning on our data, including MinerU-Diffusion~\cite{dong2026minerudiffusionrethinkingdocumentocr}, MinerU2.5-Pro~\cite{wang2026mineru25propushinglimitsdatacentric}, Dolphinv2\cite{feng2026dolphinv2universaldocumentparsing}, and Qwen2.5-VL~\cite{bai2025qwen25vltechnicalreport}.
More details on the training data in Appendix~\ref{appendix:training data} and hyperparameter settings in\ref{appendix:Hyperparameter Settings}.

\subsection{Main Results}
\label{sec:main_results}

As shown in Tab.~\ref{tab:recog}, PA-BDM achieves the best performance among diffusion-based models on most benchmarks and delivers the highest inference throughput.
Most compared models use Qwen2TokenizerFast-based tokenizers with only minor additional tokens, except for LaViDa.
Thus, TPS provides a largely comparable measure of decoding efficiency.
PA-BDM reaches 267.2 TPS, substantially outperforming DiffusionVL and MinerU-Diffusion, while maintaining peak memory comparable to other 3B-scale models.
These results show that PA-BDM improves the speed--accuracy trade-off without increasing memory cost.

The gains are more pronounced on structure-sensitive tasks such as formula, diagram, and table recognition, while improvements on plain text are more modest.
We attribute this to PA-BDM's prefix-consistent causal information flow, which better matches the strict token order and structural boundaries of LaTeX, HTML, and Mermaid outputs.
In contrast, bidirectional denoising may introduce boundary-dependent conditioning patterns, as analyzed in Sec.~\ref{sec:block_direction}.

The efficiency gain mainly comes from prefix-adaptive decoding.
By committing reliable prefixes before whole-block completion, reusing their KV states, and resetting unresolved suffixes as new candidate ranges, PA-BDM avoids decoding over shrinking residual masks and restores a large parallel decoding space.
A detailed PPC ablation is provided in Sec.~\ref{sec:ppc_ablation}.



\subsection{Model Scale Analysis}
\label{sec:model_scale}

\begin{figure}[h]
  \centering
  \includegraphics[width=1.0\linewidth, trim= 0 220 540 0, clip]{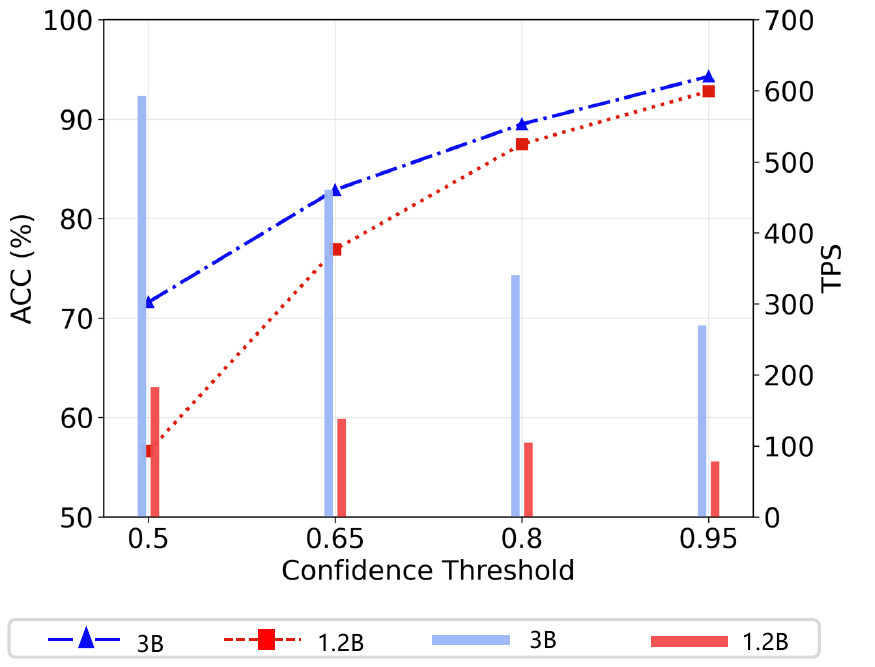}
  \caption{
Accuracy--efficiency trade-off of PA-BDM across model scales. The x-axis denotes the PPC confidence threshold.
Lines show accuracy, and bars show inference throughput (TPS).
}
\label{fig:param}
\end{figure}

We compare 1.2B and 3B PA-BDM on the English subset of OmniDoc, where layout detection is skipped and recognition is directly performed on the original images, to analyze how model scale affects recognition accuracy and inference efficiency.
As shown in Fig.~\ref{fig:param}, the 1.2B model approaches the accuracy of the 3B model under higher PPC confidence thresholds, suggesting limited marginal accuracy gains from further scaling on these recognition tasks.
However, its single-sample TPS is consistently lower across all thresholds.

We attribute this to PA-BDM's decoding dynamics.
Its efficiency depends not only on the per-forward cost, but also on how many reliable prefix tokens can be committed at each step.
Larger models tend to form longer reliable prefixes, enabling earlier KV-cache reuse and fewer redundant decoding rounds.
Thus, model scale affects both accuracy and effective decoding parallelism.
The average number of decoded tokens per forward pass is provided in Appendix~\ref{appendix:forward}.

\subsection{Ablation Study}
\label{sec:ablation}
We perform ablations on the English subsets of OmniDoc, focusing on attention design, CSL, and PPC.
Additional hyperparameter studies are provided in the Appendix \ref{appendix:supp}.

\begin{table}[h]
\centering
\small
\setlength{\tabcolsep}{4pt}
\begin{adjustbox}{max width=\columnwidth}
\begin{tabular}{c|cc|cc|cc}
\toprule
\multirow{2}{*}{Block Size}
& \multicolumn{2}{c|}{Formula $\uparrow$} 
& \multicolumn{2}{c|}{Text $\downarrow$} 
& \multicolumn{2}{c}{Table $\uparrow$} \\
\cmidrule(lr){2-3}\cmidrule(lr){4-5}\cmidrule(lr){6-7}
& Bidir. & Causal & Bidir. & Causal & Bidir. & Causal \\
\midrule
8  & 76.2 & 87.1 & 0.214 & 0.197 & 75.3 & 83.5 \\
16 & 69.2 & 78.0 & 0.223 & 0.226 & 61.7 & 74.2 \\
32 & 31.4 & 27.5 & 0.271 & 0.254 & 38.7 & 45.2 \\
\bottomrule
\end{tabular}
\end{adjustbox}
\caption{
Effect of intra-block attention direction under different block sizes.
Bidir. and Causal denote bidirectional and causal intra-block attention, respectively.
All variants are trained without CSL or PPC and evaluated with one-shot decoding.
}
\label{tab:block_size_direction}
\end{table}


\subsubsection{Effect of Intra-block Modeling Direction}
\label{sec:block_direction}
To isolate the effect of intra-block information flow, we remove CSL and PPC, change only the intra-block attention direction, and evaluate all models with vanilla one-shot decoding.

\begin{figure}[t]
  \centering
  \includegraphics[width=0.9\linewidth, trim= 0 120 500 0, clip]{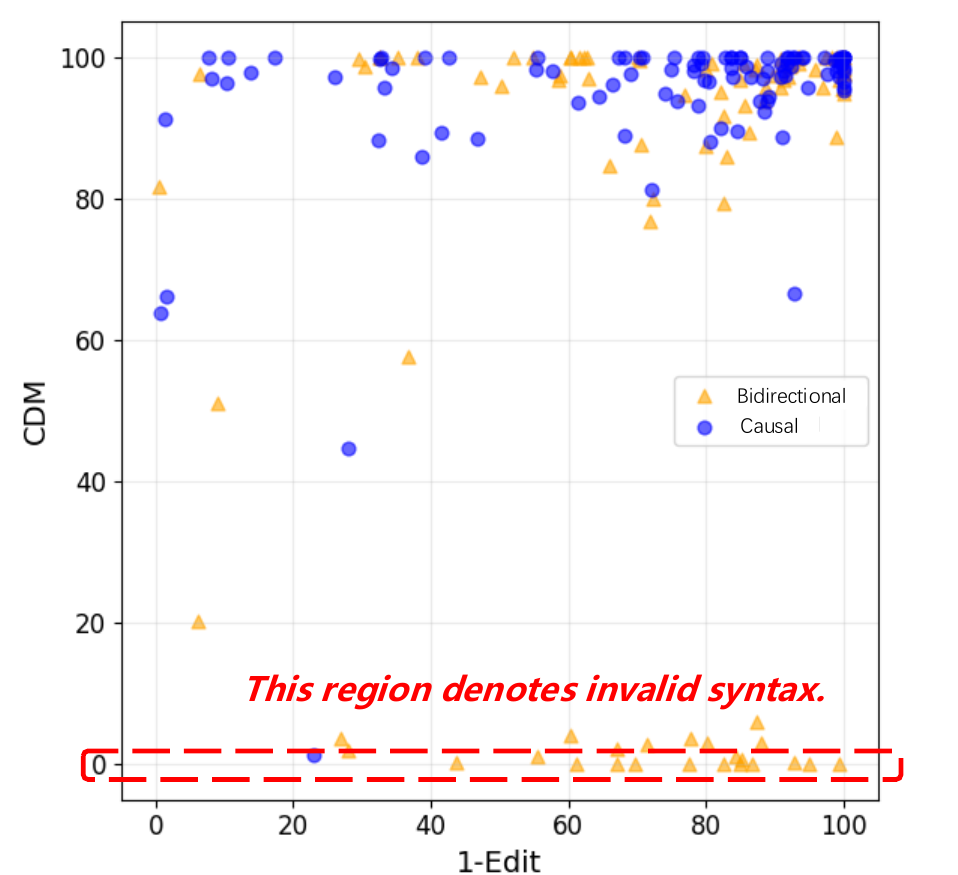}
  \caption{
\textbf{Sample-level comparison between bidirectional and causal intra-block modeling on 100 formula instances.}
Each point denotes one sample under normalized $1-\mathrm{Edit}$ and CDM~\cite{wang2024cdmreliablemetricfair}.
}
\label{fig:cdm_vs_edit}
\end{figure}

Table~\ref{tab:block_size_direction} shows that causal intra-block modeling performs better than bidirectional modeling in most settings, especially on formula and table recognition.
Although causal modeling is not uniformly better in every case, the results suggest that additional bidirectional context is not always beneficial for structured discrete sequences.
A prefix-to-suffix dependency path that is consistent with inter-block autoregressive progression may better match token order and structural boundary modeling.
As the block size increases, both variants degrade, indicating that fixed-range one-shot prediction is itself unstable for longer structured sequences.
Fig.~\ref{fig:cdm_vs_edit} further illustrates this difference at the sample level.
The two variants appear similar under normalized $1-\mathrm{Edit}$, suggesting that their surface-level character similarity can be close.
However, CDM depends on successful formula rendering and character matching in the rendered space, so syntactically invalid predictions tend to receive near-zero scores.
The bidirectional variant produces more near-zero CDM samples, indicating that character-level similarity does not necessarily imply structural validity.
This suggests that bidirectional intra-block context may lead to more structural failures in some cases.


\begin{table}[htbp]
\centering
\small
\setlength{\tabcolsep}{3.5pt}
\begin{tabular*}{\columnwidth}{@{\extracolsep{\fill}}l|cc|cc|cc}
\toprule
\multirow{2}{*}{Objective} 
& \multicolumn{2}{c|}{60K steps} 
& \multicolumn{2}{c|}{120K steps} 
& \multicolumn{2}{c}{180K steps} \\
\cmidrule(lr){2-3}\cmidrule(lr){4-5}\cmidrule(lr){6-7}
& ACC $\uparrow$ & TPS $\uparrow$ 
& ACC $\uparrow$ & TPS $\uparrow$ 
& ACC $\uparrow$ & TPS $\uparrow$ \\
\midrule
CE     & 56.1 & 34.5  & 76.7 & 89.6  & 82.7 & 113.2 \\
Random & 48.7 & 29.7  & 67.0 & 45.8  & 75.6 & 61.0  \\
CSL    & \textbf{81.2} & \textbf{147.5} 
       & \textbf{94.0} & \textbf{246.1} 
       & \textbf{94.1} & \textbf{251.0} \\
\bottomrule
\end{tabular*}
\caption{
Effect of CSL.
All variants use the same PA-BDM inference setting and differ only in the training objective.
Random drops supervised positions randomly instead of using prefix confidence.
}
\label{tab:csl_ablation}
\end{table}

\subsubsection{Effect of the CSL Objective}
\label{sec:csl_ablation}

We evaluate CSL under the same PA-BDM decoding setting.
Standard CE supervises all masked suffix positions uniformly.
However, under causal intra-block denoising, later masked tokens may depend on earlier uncertain masked tokens, making full-suffix supervision noisy.
CSL instead selects the supervised range according to prefix confidence, keeps the first low-confidence token as the learning frontier, and excludes positions to its right.

As shown in Table~\ref{tab:csl_ablation}, CSL consistently improves ACC across training steps.
The Random baseline performs much worse than both CE and CSL, indicating that the gain comes from prefix-aware supervision rather than simply reducing the number of supervised tokens.
Since all variants use the same inference algorithm, the higher TPS of CSL suggests that it learns longer reliable prefixes instead of only the leftmost positions.
Otherwise, PPC would commit fewer tokens per forward pass and require more decoding rounds.
The simultaneous improvement in ACC and TPS indicates that CSL progressively pushes the reliable frontier to the right during training.

Threshold sensitivity analysis is provided in Appendix~\ref{appendix:csl_threshold}, showing that CSL remains stable within a reasonable range of confidence thresholds.

\begin{figure}[tbp]
  \centering
  \includegraphics[width=1.0\linewidth, trim= 0 0 0 0, clip]{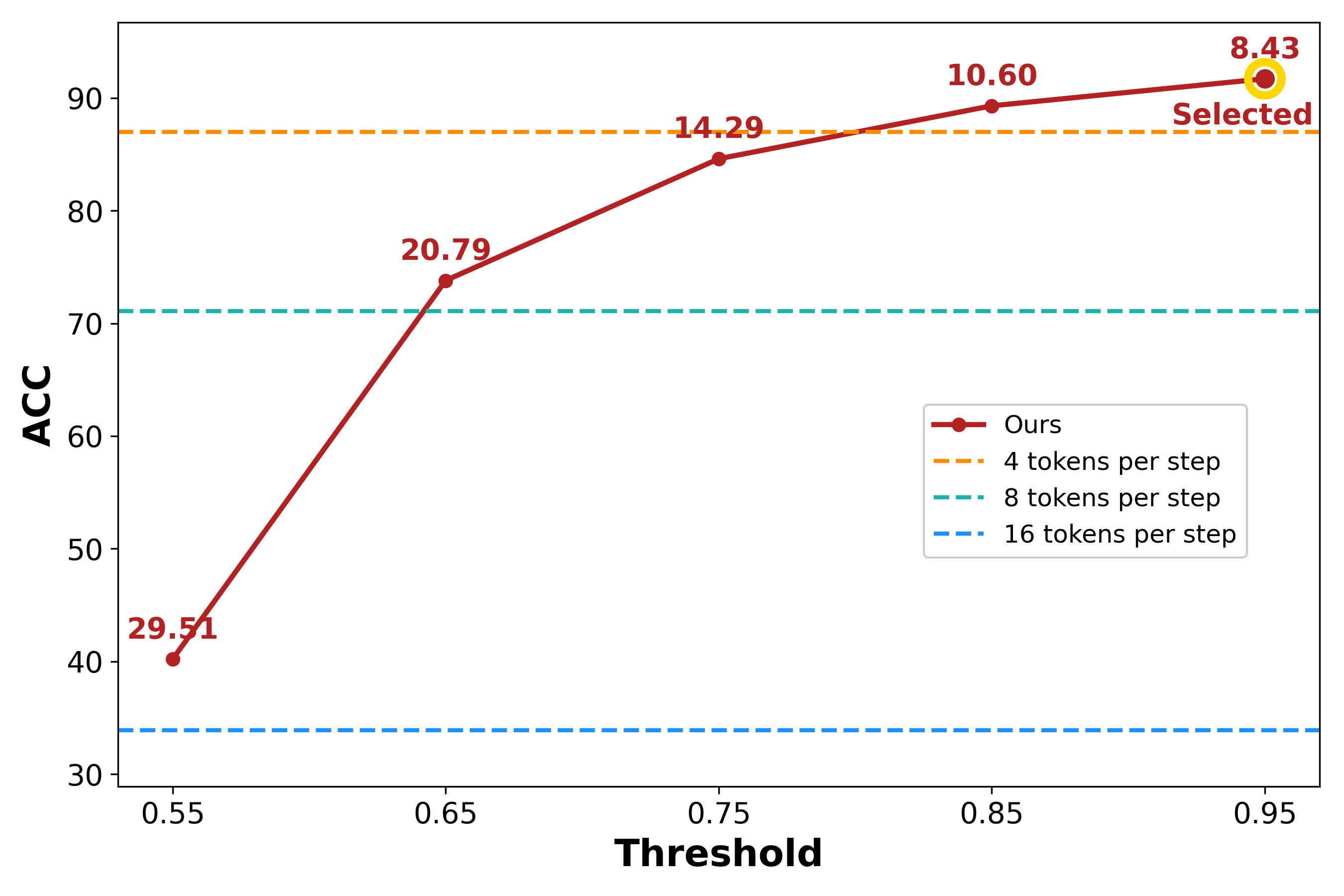}
  \caption{
The red line shows the ACC across different confidence thresholds.
Numbers along the red line indicate the average number of tokens decoded at each step.
The three dashed lines represent the accuracy of the baseline method when predicting a fixed number of 4, 8, or 16 tokens at each step.
}
  \label{fig:threshold}
\end{figure}

\subsubsection{Effect of PPC Decoding}
\label{sec:ppc_ablation}

We compare PPC with fixed commit strategies.
A fixed strategy commits a preset number of leftmost tokens at each forward pass, whereas PPC selects the commit length adaptively based on prefix confidence.
As shown in Fig.~\ref{fig:threshold}, fixed strategies provide only a limited accuracy--parallelism trade-off.
A small fixed length is conservative and underuses parallel decoding, while a large fixed length may commit unstable tokens in structurally difficult regions.
PPC avoids this fixed-granularity constraint by adjusting the commit length according to the confidence profile of each candidate range.
As the confidence threshold increases, PPC commits fewer tokens per forward and becomes more conservative, showing that confidence effectively controls how much prefix information is selected for commitment.
At the default threshold of 0.95, PPC commits 8.4 tokens per forward on average, much longer than the fixed 4-token strategy, while still achieving higher ACC.
This suggests that the reliable prefix length varies across decoding states, and confidence-based adaptive commitment is more effective than a fixed generation granularity.
The case study in Fig.~\ref{fig:study_case} further illustrates this behavior.

We further isolate the efficiency components of PPC in Table~\ref{tab:efficiency_components}.
The baseline still uses KV cache, but only after the whole block is completed.
Reset improves TPS by refreshing the unresolved suffix as a new candidate window, while Prefix-level Cache avoids recomputing already committed tokens.
Both components independently improve throughput to different extents.
Combining them achieves the highest TPS, increasing from 124.9 to 269.1, with nearly unchanged accuracy.

\begin{table}[t]
\centering
\small
\begin{tabular*}{\columnwidth}{@{\extracolsep{\fill}}l|cc|cc}
\toprule
Variant & Reset & Prefix Cache & TPS $\uparrow$ & ACC $\uparrow$ \\
\midrule
Block-level &            &            & 124.9 & \textbf{94.5} \\
+ Prefix Cache &         & \checkmark & 227.7 & 94.5 \\
+ Reset        & \checkmark &         & 193.6 & 94.3 \\
Full           & \checkmark & \checkmark & \textbf{269.1} & 94.3 \\
\bottomrule
\end{tabular*}
\caption{
Ablation on the efficiency components of PPC.
Block-level denotes the default strategy that caches tokens only after the entire block is completed.
Reset denotes resetting the unresolved suffix as a new candidate decoding window after prefix commitment.
Prefix-level Cache denotes immediately caching committed prefix tokens before the entire block is completed.
}
\label{tab:efficiency_components}
\end{table}



\section{Conclusion}

In this paper, we revisit block diffusion models for document parsing and identify two practical limitations of standard BDMs, namely fixed block-level granularity and inconsistent information flow between inter-block autoregression and intra-block bidirectional denoising. To address these issues, we propose PA-BDM, which treats each block as a maximum candidate generation range rather than a fixed commitment unit. By combining causal intra-block modeling, PPC decoding, and CSL training, PA-BDM enables earlier reliable-prefix commitment, timely KV-cache reuse, and parallel-space resetting. Experiments on text, formula, table, and diagram recognition show that PA-BDM substantially improves inference throughput while achieving a better speed--accuracy trade-off than comparable diffusion-based baselines, especially on structure-sensitive tasks. These results suggest reliable-prefix modeling as an effective direction for improving the efficiency of diffusion-based document recognition.

\section*{Limitations}

Our experiments are mainly conducted on public datasets that provide both training and test splits.
These datasets enable controlled training and fair comparison, but they also limit the diversity of evaluation scenarios.
Most available open-source training data are still English-centric and cover a relatively limited range of document styles, languages, and layout structures.
Meanwhile, several recent and more challenging benchmarks only provide test sets, making it difficult to conduct fully controlled evaluation under the same training setting.
Therefore, we have not systematically evaluated PA-BDM in more complex multilingual scenarios, such as Chinese formulas, Chinese tables, multilingual mixed documents, and more complicated page-level layouts.
These scenarios usually involve richer character sets, different formatting conventions, and more complex structural dependencies, which may pose additional challenges to reliable-prefix modeling and structured generation.
As a result, the current experiments mainly validate the effectiveness of PA-BDM under relatively standard open-source settings.
Its generalization to broader languages, layouts, and real-world document scenarios remains to be further studied.

\section*{Acknowledgments}
The authors wish to thank the anonymous reviewers for their helpful comments and WisPaper for its assistance in literature retrieval. This work was partially funded by National Key R\&D Program of China No.2025ZD0215702, National Natural Science Foundation of China (No. 62576106, 62476061, 62376061).


\bibliography{custom}

\clearpage
\appendix

\section{Evaluation Metrics}
\label{appendix:metrics}

In the main experiments, we evaluate PA-BDM on text, formula, table, and diagram recognition tasks.
For text recognition, we use Edit Distance (Edit), which measures the surface-level character difference between the predicted text and the ground truth.
However, surface-level character similarity is less suitable for structured outputs such as formulas and tables, where small token changes may lead to different structures and different textual forms may represent similar semantics.
Therefore, we use task-specific metrics for these structured recognition tasks.

For formula recognition, we report Character Detection Matching (CDM)~\cite{wang2024cdmreliablemetricfair}, which evaluates the matching quality of rendered mathematical expressions at the character level.
For table recognition, we report Tree-Edit-Distance-based Similarity (TEDS)~\cite{zhong2020imagebasedtablerecognitiondata}, which measures the similarity between predicted and ground-truth HTML table trees while accounting for both table structure and cell content.
For diagram recognition, we report F1 based on the parsed Mermaid graphs, where nodes and edges are extracted from the generated and reference Mermaid code and then matched for evaluation.

Several ablation studies in the main text are conducted on the English subsets of OmniDoc~\cite{ouyang2025omnidocbenchbenchmarkingdiversepdf}, covering text, table, and formula recognition.
When reporting a single aggregate ACC score, we compute it as the average of normalized text accuracy, table TEDS, and formula CDM:
\begin{equation*}
\resizebox{\columnwidth}{!}{$
\text{ACC} =
\frac{(1 - \text{Text}^{\text{Edit}}) \times 100 + \text{Table}^{\text{TEDS}} + \text{Formula}^{\text{CDM}}}{3}.
$}
\end{equation*}
Here, $\text{Text}^{\text{Edit}}$ denotes the edit-distance score for text recognition, while $\text{Table}^{\text{TEDS}}$ and $\text{Formula}^{\text{CDM}}$ denote the corresponding table and formula scores.
Higher ACC indicates better overall recognition performance across these three task types.

\section{Training Details}

\subsection{Training Data}
\label{appendix:training data}

The training data for table recognition, formula recognition, and diagram recognition are directly taken from the training splits of the corresponding evaluation benchmarks described in the main text. For text recognition, we combine existing document-layout annotations with an additional scientific-document-oriented corpus to improve coverage across general and academic document scenarios.

Our text recognition data are built from two sources. The first source is DocLayNet, which provides high-quality human annotations for document layout elements across diverse domains. We use its annotated text regions to obtain general document text samples. 

To better support scientific document recognition, we further construct an element-level paragraph dataset from the Text\_Completion\_arXiv corpus. This corpus provides page-level annotations for arXiv papers. Based on these annotations, we derive block-level image--text pairs through an automatic filtering and alignment pipeline. Specifically, we first apply an existing layout detection model to split each page image into layout elements and crop the corresponding regions according to the detected bounding boxes. We then use an existing strong text recognition model to generate candidate transcriptions for each cropped region.

In parallel, we extract paragraph candidates from the original page-level text by using \texttt{\textbackslash n\textbackslash n} as paragraph boundaries. The recognized texts and the extracted paragraph candidates are then matched under document reading-order constraints. We use edit-distance-based filtering with strict thresholds to remove unreliable pairs and retain only high-confidence alignments.

We emphasize that this data construction pipeline is not intended as a methodological contribution of this paper. It mainly relies on existing datasets, existing layout and recognition tools, and engineering-level filtering and alignment procedures. Its purpose is to provide cleaner and broader text-recognition supervision for training PA-BDM under document parsing scenarios. For reproducibility, we will release the constructed training data, filtering scripts, and data processing pipeline together with the training and inference code.

The resulting text recognition data contain both general document elements from DocLayNet and scientific paragraph samples derived from arXiv papers. This design provides broader supervision for text recognition and improves the robustness of the model in scenarios involving dense paragraphs, inline formulas, and academic writing styles.

\subsection{Training Configuration}
\label{appendix:training_configuration}

We train PA-BDM at two model scales, 1.2B and 3B. All models are trained for 210K optimization steps on 4 NVIDIA H100 80GB GPUs, with a per-GPU batch size of 6 and gradient accumulation over 5 steps, yielding an effective global batch size of 120. We use AdamW with a learning rate linearly warmed up from $1\times10^{-8}$ to $5\times10^{-5}$, followed by cosine decay. Unless otherwise specified, the maximum candidate block size is set to 32, and the confidence threshold for both CSL and PPC-based prefix caching is set to 0.95. Speed is measured on a single NVIDIA RTX 4090 GPU with batch size 1.



\begin{table*}[ht]
\centering
\small
\setlength{\tabcolsep}{10pt}
\begin{tabular}{lccccc}
\toprule
Size & $\tau$ & Peak Mem $\downarrow$ & 
Avg. tokens / forward $\uparrow$ & Forward calls $\downarrow$ & 
Forward time $\downarrow$ \\
\midrule
1.2B & 0.65 & 2.3 & 3.6 & 65.2 & 0.026 \\
1.2B & 0.80 & 2.3 & 2.4 & 94.1 & 0.024 \\
1.2B & 0.95 & 2.3 & 1.8 & 125.6 & 0.023 \\
\midrule
3B & 0.65 & 7.4 & 14.3 & 15.7 & 0.033 \\
3B & 0.80 & 7.3 & 10.6 & 21.3 & 0.031 \\
3B & 0.95 & 7.3 & 8.4 & 27.2 & 0.031 \\
\bottomrule
\end{tabular}
\caption{
Decoding statistics of PA-BDM across model scales and PPC confidence thresholds on the English subset of OmniDoc.
Although the smaller model has a lower per-forward cost, the larger model tends to commit more tokens per forward pass, reducing the number of decoding rounds and improving overall throughput.
}
\label{tab:forward_stats}
\end{table*}

\section{Hyperparameter Settings}
\label{appendix:Hyperparameter Settings}

For a fair comparison, we follow the official configurations of each baseline whenever available. For diffusion-based visual language models, we use the same candidate block length and confidence threshold unless otherwise specified. Specifically, the block length is set to 32, and the confidence threshold is set to 0.95.

\noindent
\textbf{LaViDa.}
LaViDa\cite{li2025lavidalargediffusionlanguage} is a representative diffusion-based visual language model. In standard diffusion visual language models with global bidirectional attention, the hidden states of previously decoded tokens may change when new tokens are generated. As a result, these models cannot directly support exact KV cache reuse in the same way as autoregressive models. Although several cache mechanisms have been proposed for diffusion language models, they usually rely on approximation and may introduce a mismatch between training and inference.

This issue may have a limited effect on semantic-level understanding tasks, but it can be more problematic for token-level recognition tasks, especially when the output contains strict syntactic structures such as LaTeX formulas or HTML tables. LaViDa addresses part of this issue with Prefix KV, where the visual prefix is excluded from bidirectional denoising and can therefore be cached. We consider this design reasonable because visual tokens should remain independent of the generated textual content, and visual tokens often account for a large portion of the computation. Compared with more aggressive cache strategies, Prefix KV caches fewer tokens, but it better preserves consistency between training and inference.


In our main experiments, we keep the Prefix KV strategy used by LaViDa. Following common diffusion decoding practice, LaViDa performs iterative denoising within a fixed generation space.

\noindent
\textbf{DiffusionVL.}
DiffusionVL\cite{zeng2026diffusionvltranslatingautoregressivemodels} converts Qwen2.5-VL\cite{bai2025qwen25vltechnicalreport} from an autoregressive visual language model into a block diffusion visual language model. Our PA-BDM is built upon the same backbone, but differs in the attention mechanism and decoding strategy. DiffusionVL follows the standard BDM formulation, where each block is generated as a fixed unit and cached only after the whole block is completed. We use its default block-wise generation and cache update strategy. The block length is set to 32 during both training and decoding, and the confidence threshold is set to 0.95.

\noindent
\textbf{MinerU-Diffusion.}
MinerU-Diffusion\cite{dong2026minerudiffusionrethinkingdocumentocr} is mainly designed for document parsing. Unlike many related works that mainly focus on text recognition, it reports detailed results on structured recognition tasks such as formula and table recognition, making it a useful baseline for our evaluation. We follow the default configuration released in its official codebase. The block length is set to 32, and the confidence threshold is set to 0.95.
\section{Supplement Experiments}
\label{appendix:supp}

\subsection{Model Scale and Decoding Dynamics}
\label{appendix:forward}

Existing ARM-based document parsing models can often achieve comparable or even better recognition accuracy with around 1B parameters than larger 3B models~\cite{niu2025mineru25decoupledvisionlanguagemodel,cui2025paddleocrvlboostingmultilingualdocument,duan2026glmocrtechnicalreport}.
This suggests that, for recognition tasks in document parsing, further scaling may provide limited marginal accuracy gains while increasing per-step computation and memory cost.

However, PA-BDM exhibits a different scaling behavior due to its prefix-adaptive decoding dynamics.
Its inference efficiency depends not only on the cost of each forward pass, but also on how many reliable prefix tokens can be committed at each step.
Larger models tend to form high-confidence prefixes more quickly, allowing PPC to commit more tokens per forward pass and reduce the total number of forward calls.
As a result, although the 3B model has higher per-forward latency and memory usage, it can still achieve higher throughput than the 1.2B model in the batch-size-one setting.

Table~\ref{tab:forward_stats} further illustrates this effect.
At $\tau=0.95$, the 1.2B model commits only 1.8 tokens per forward pass on average and requires 125.6 forward calls, while the 3B model commits 8.4 tokens per forward pass and requires only 27.2 forward calls.
Based on the average committed tokens per forward pass and the average forward time, the estimated forward-level TPS of the 3B model is about 3.46$\times$ that of the 1.2B model.
Across different confidence thresholds, this ratio is around 3.1--3.5$\times$.

Considering memory usage, the peak memory of the 3B model is about 3.2$\times$ that of the 1.2B model, while its estimated forward-level TPS is about 3.1--3.5$\times$ higher across thresholds.
This indicates that, especially under higher confidence thresholds, the faster reliable-prefix growth of the 3B model can largely compensate for its higher memory cost.
Therefore, for PA-BDM, a smaller model does not necessarily lead to higher practical throughput, since model scale affects both computation cost and effective decoding parallelism.

\begin{table}[t]
\centering
\small
\setlength{\tabcolsep}{4pt}
\resizebox{\columnwidth}{!}{
\begin{tabular}{lcccccc}
\toprule
\multirow{2}{*}{$\tau$}
& \multicolumn{2}{c}{60K}
& \multicolumn{2}{c}{120K}
& \multicolumn{2}{c}{180K} \\
\cmidrule(lr){2-3}
\cmidrule(lr){4-5}
\cmidrule(lr){6-7}
& ACC & Ratio
& ACC & Ratio
& ACC & Ratio \\
\midrule
0.65 & 84.7 & 0.92 & 85.1 & 0.96 & 86.7 & 0.96 \\
0.80 & 85.4 & 0.59 & 90.1 & 0.71 & 92.2 & 0.77 \\
0.95 & 81.2 & 0.31 & 94.0 & 0.57 & 94.1 & 0.63 \\
\bottomrule
\end{tabular}
}
\caption{
Sensitivity analysis of the CSL confidence threshold $\tau$ across training steps.
ACC is evaluated under the same PA-BDM decoding setting, with the inference confidence threshold fixed to 0.95.
Ratio denotes the ratio of actually supervised tokens to valid masked tokens under CSL, excluding padding positions.
For the 60K, 120K, and 180K columns, Ratio is averaged over the training intervals 0--60K, 60--120K, and 120--180K, respectively.
All variants use the same random seed for suffix sampling.
}
\label{tab:csl_threshold_steps}
\end{table}

\subsection{Sensitivity to CSL Confidence Threshold}
\label{appendix:csl_threshold}

We further analyze the effect of the CSL confidence threshold $\eta$.
This threshold controls how conservatively CSL expands supervision to longer continuations.
A smaller $\eta$ supervises more masked tokens, while a larger $\eta$ requires a more reliable prefix before including right-side positions in the loss.

As shown in Table~\ref{tab:csl_threshold_steps}, different thresholds lead to different supervised ratios.
When $\eta=0.65$, the supervised ratio reaches $0.92$ during the first 60K steps, providing dense supervision and relatively strong early accuracy.
In contrast, $\eta=0.95$ is more conservative at the beginning, supervising only $0.31$ of valid masked tokens and therefore showing lower early accuracy.

Across all settings, the supervised ratio increases as training progresses.
For example, it increases from $0.31$ to $0.63$ when $\eta=0.95$, and from $0.59$ to $0.77$ when $\eta=0.80$.
This suggests that CSL does not collapse to learning only the leftmost positions.
Instead, as earlier masked tokens become more reliable, the learning frontier gradually moves to the right and more continuation tokens enter the supervised range.

The final accuracy shows that denser supervision is not always better.
Although $\eta=0.65$ supervises more tokens throughout training, its final ACC is only $86.7$, while $\eta=0.95$ reaches $94.1$ with a lower supervised ratio.
We attribute this to causal intra-block denoising, where right-side masked tokens may depend on uncertain high-entropy tokens on their left.
Supervising these right-side positions too early can introduce noisy gradients from unstable prefix states.
A higher threshold avoids such noisy continuation supervision by expanding the supervised range only after the prefix becomes sufficiently reliable.

\begin{table}[t]
\centering
\small
\setlength{\tabcolsep}{6pt}
\begin{tabular}{lccc}
\toprule
$D$ & ACC $\uparrow$ & Ratio $\uparrow$ & Avg. tokens / forward $\uparrow$ \\
\midrule
8  & 94.2 & 0.97 & 6.1 \\
16 & 94.1 & 0.93 & 7.6 \\
32 & 94.1 & 0.71 & 8.4 \\
64 & 92.8 & 0.43 & 8.1 \\
\bottomrule
\end{tabular}
\caption{
Sensitivity analysis of the maximum candidate block size $D$.
ACC measures recognition performance.
Ratio denotes the proportion of actually supervised tokens among valid masked tokens under CSL, measured after the training loss becomes relatively stable.
Avg. tokens / forward denotes the average number of committed tokens per forward pass under PPC.
}
\label{tab:block_size_sensitivity}
\end{table}

\subsection{Sensitivity to Maximum Candidate Block Size}
\label{appendix_block_size}

Table~\ref{tab:block_size_sensitivity} analyzes the effect of the maximum candidate block size $D$.
Since PA-BDM samples a suffix start and masks the positions to its right, smaller blocks usually provide a shorter valid masked suffix.
As a result, when $D=8$ or $D=16$, CSL can cover almost the entire masked suffix after training becomes stable, leading to high supervised ratios of $0.97$ and $0.93$.
However, a higher Ratio under small block sizes does not necessarily indicate stronger long-range prefix learning.
It partly results from the shorter masked suffix, which makes the valid supervision range easier to cover.
Meanwhile, small blocks also cap the candidate range available to PPC.
Although $D=8$ and $D=16$ achieve strong ACC, their average committed tokens per forward pass remain lower, indicating that their inference parallelism is limited by the small candidate range.

Increasing $D$ to $32$ provides more decoding headroom while still maintaining sufficient CSL supervision.
In practice, the committed length is usually concentrated in a moderate range, such as 4--12 tokens per forward pass.
However, for easier local regions, PPC can occasionally commit more than 20 or even close to 30 tokens in one step.
Therefore, a sufficiently large candidate range is still useful, even when the average committed length is much smaller than the block size.
This explains why $D=32$ can achieve the highest average committed length while maintaining comparable ACC.

In contrast, further increasing $D$ to $64$ reduces both ACC and average committed length.
This shows that enlarging the candidate range does not automatically improve effective parallelism.
The actual committed length is still determined by the model's ability to form reliable prefixes, so the block size serves as an upper bound rather than the expected decoding length.
When $D$ is excessively large, most additional candidate positions are rarely committed, while the number of blocks per sequence decreases under a fixed sequence length.
Since CSL only supervises tokens up to the reliable frontier in each block, this can make the overall supervision signal sparser and training less stable.
Moreover, much of the enlarged candidate range remains unused while introducing more uncertain masked context during training.

Therefore, a moderate block size is preferable.
It provides enough headroom for occasional long reliable-prefix commitment, avoids the ceiling effect of overly small blocks, and prevents the supervision sparsity caused by overly large blocks.
These results support our default choice of $D=32$.

\begin{table}[t]
  \centering
  \setlength{\tabcolsep}{3pt}
  \resizebox{\columnwidth}{!}{
  \begin{tabular}{lc|cccc|cc}  
  \toprule
    \textbf{Model} & \textbf{Size} 
    & Precision$\uparrow$ & Recall$\uparrow$ & F1$\uparrow$ & FPS$\uparrow$
    & \textbf{\st{NMS}} & \textbf{\st{Conf}} \\
    \midrule
    \multicolumn{8}{c}{\textit{Vision Models}}\\
    \midrule
    YOLOv11m & 20M & 87.5 & 95.4 & 91.3 & 21.3 & \redXSolidBrush & \redXSolidBrush \\
    Doc-YOLO & 20M & 88.0 & 96.3 & 91.9 & 15.7 & \greenCheckmarkBold & \redXSolidBrush \\
    \midrule
    \multicolumn{8}{c}{\textit{Pipeline Tool}}\\
    \midrule
    PP-StructureV3 & - & 91.4 & 94.7 & 93.0 & 14.1 & - & - \\
    \midrule
    \multicolumn{8}{c}{\textit{Vision-Language Models}}\\
    \midrule
    Qwen2.5-VL &  3B & 91.3 & 95.7 & 93.5 & 1.3 & \greenCheckmarkBold & \greenCheckmarkBold \\
    PA-BDM & 3B & 91.8 & 95.9 & 93.8 & 5.7 & \greenCheckmarkBold & \greenCheckmarkBold \\
    \bottomrule
  \end{tabular}
  }
  \caption{
  Comparison of layout detection methods.
  \textbf{\st{NMS}} and \textbf{\st{Conf}} respectively indicate that Non-Maximum Suppression and confidence adjustment are not required.
  \textit{For FPS evaluation, the batch size is set to 1, which relatively reduces the advantage of models with fewer parameters.}
  }
  \label{table:dla}
\end{table}

\subsection{Layout Detection}
\label{Layout Detection}

VLM-based object detection can be viewed as a token-level recognition task, where the model is required to produce precise and unambiguous structured predictions.
Recent document parsing systems increasingly unify text recognition and layout element detection within autoregressive VLMs, which simplifies the overall system design.
However, a major limitation of this paradigm is that autoregressive models generate outputs token by token, leading to much lower inference efficiency than parallel vision-based detectors.

We evaluate PA-BDM on DocLayNet~\cite{Pfitzmann_2022} to examine the feasibility of applying diffusion-based decoding to structured document element recognition within a unified VLM framework.
As shown in Table~\ref{table:dla}, vision-based detectors still achieve higher FPS, but VLM-based methods benefit from richer multimodal representations and obtain better detection accuracy.
Compared with single-token decoding, PA-BDM improves inference efficiency by generating multiple tokens in parallel.
For example, decoding five tokens per iteration achieves more than a 4$\times$ speedup over single-token decoding, while causing only a small F1 drop of 0.4 points.

From a practical perspective, vision-based detectors usually require additional post-processing steps such as Non-Maximum Suppression and confidence adjustment, which increases system complexity.
In contrast, VLM-based methods can directly generate structurally valid outputs, reducing the need for task-specific post-processing and making the unified document parsing pipeline simpler and more robust.

\begin{table}[t]
\centering
\small
\setlength{\tabcolsep}{8pt}
\begin{tabular}{lcc}
\toprule
\textbf{Method} & \textbf{Size} & \textbf{OmniDocBench} \\
\midrule
\multicolumn{3}{l}{\textit{Specialized OCR}} \\
dots.ocr        & 3B   & \textbf{0.032} \\
DeepSeek-OCR    & 3.4B & 0.049 \\
MinerU 2.0 VLM  & 0.9B & 0.045 \\
MonkeyOCR-pro   & 3B   & 0.058 \\
Mistral OCR     & -    & 0.072 \\
olmOCR          & 7B   & 0.097 \\
Nanonets-OCR-s  & 3B   & 0.134 \\
SmolDocling     & 256M & 0.262 \\
\midrule
\multicolumn{3}{l}{\textit{Autoregressive VLMs}} \\
Qwen 2.5 VL     & 72B  & 0.092 \\
Qwen 2.5 VL     & 7B   & 0.135 \\
Qwen 2.5 VL     & 3B   & 0.184 \\
\midrule
\multicolumn{3}{l}{\textit{Diffusion VLMs}} \\
Dimple          & 7B   & 0.856 \\
LaViDa-L        & 8B   & 0.994 \\
LLaDA-V         & 7B   & 0.524 \\
DODO        & 3B & 0.066 \\
DODO \textit{fast} & 3B & 0.159 \\
\midrule
\multicolumn{3}{l}{\textit{Ours}} \\
\textbf{PA-BDM}    & 3B & 0.061 \\
\bottomrule
\end{tabular}
\caption{
Page-level OCR comparison on the English subset of OmniDocBench under the DODO evaluation setting.
Results of external baselines are taken from DODO.
Lower normalized edit distance is better.
}
\label{tab:omnidocbench_ocr}
\end{table}

\subsection{Page-level Parsing Evaluation}
\label{appendix:page_level}

Document parsing evaluation has gradually moved from isolated component-level evaluation to page-level end-to-end evaluation.
In this setting, a system can either first detect layout elements and then recognize each cropped region, or directly take the whole page as input and generate the complete page-level transcription.
This evaluation protocol better reflects practical document parsing scenarios, but it also introduces a larger distribution gap.
Page-level benchmarks often contain diverse layouts, non-English elements, formulas, and tables, while most publicly available training data provide only limited coverage of such diverse structures and languages.
For this reason, the main experiments in this paper focus on the English component-level subsets of OmniDocBench, where layout detection is skipped and recognition models are evaluated on more controlled input regions.

To further examine the applicability of PA-BDM to page-level parsing, we additionally follow the evaluation setting of DODO.
Specifically, we evaluate on the English subset of OmniDocBench and use Normalized Edit Distance as the page-level recognition metric.
This setting allows us to compare PA-BDM with recent specialized OCR systems, autoregressive VLMs\cite{wei2026deepseekocr2visualcausal,li2025dotsocrmultilingualdocumentlayout,Nanonets-OCR-S,poznanski2025olmocrunlockingtrillionstokens}, and diffusion-based VLMs under the same page-level protocol.

As shown in Table~\ref{tab:omnidocbench_ocr}, existing full-sequence diffusion VLMs such as Dimple\cite{yu2025dimplediscretediffusionmultimodal}, LaViDa-L\cite{li2025lavidalargediffusionlanguage}, and LLaDA-V\cite{you2025lladavlargelanguagediffusion} perform poorly on dense page-level document transcription.
This is consistent with the observation in DODO that global masked diffusion can suffer from severe alignment and structural instability on OCR-like deterministic generation tasks.
By contrast, block-based diffusion methods substantially improve the viability of diffusion decoding for page-level OCR.
DODO achieves a normalized edit distance of 0.066 on OmniDocBench, outperforming the Qwen2.5-VL autoregressive backbones of different scales and greatly improving over prior diffusion VLMs.

These results suggest that page-level document parsing is a challenging but important setting for diffusion-based VLMs.
Compared with component-level recognition, page-level parsing requires the model to jointly handle reading order, layout structure, dense text, tables, and formulas, making alignment stability especially critical.
The strong gap between full-sequence diffusion VLMs and block-based diffusion models further supports the need for constrained and prefix-consistent generation mechanisms.
PA-BDM follows this direction by treating each block as a maximum candidate range and using reliable-prefix commitment to improve both structural stability and decoding efficiency.

\section{Batch-parallel PPC Decoding}
\label{appendix:parallel_ppc}

Confidence-based block diffusion decoding methods commonly face a practical batching issue.
Since the number of committed tokens is determined by confidence, different samples may commit different numbers of tokens at each decoding step.
As a result, samples in the same batch can have different prefix lengths and may require different numbers of decoding rounds for the same relative block position, which reduces batch-level parallelism.
PPC also has this adaptive-length property, since each sample independently commits the longest reliable prefix from its candidate range.

\begin{figure}[tbp]
  \centering
  \includegraphics[width=1\linewidth, trim= 0 285 600 0, clip]{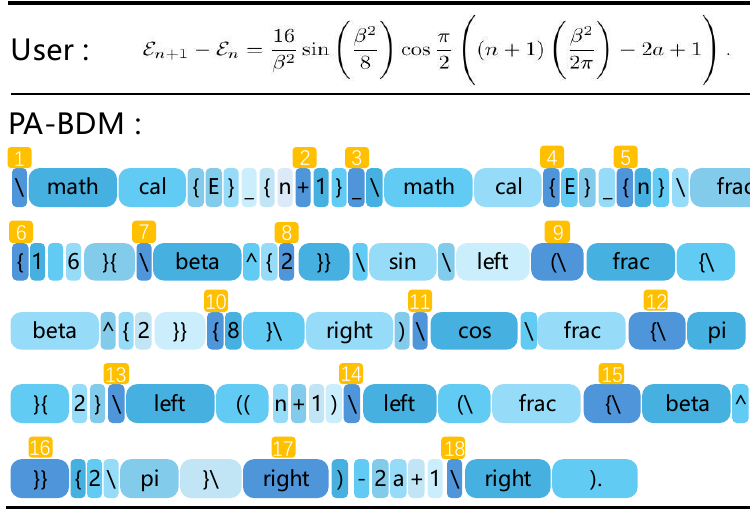}
  \vskip-.1cm 
  \caption{A case study of PA-BDM on mathematical formula recognition using adaptive step-size decoding. The number in the top-left corner of each slot indicates the generation order, while the color intensity within each slot represents the generation time (darker indicates earlier).}
  \label{fig:study_case}
\end{figure}

To preserve batch-level parallelism, we use a batch-aligned candidate construction strategy.
Let $p_i^{(r)}$ denote the current committed prefix length of the $i$-th active sample at decoding round $r$, and let $D$ be the maximum candidate block size.
For a length-bucketed batch where $\max_i p_i^{(r)}-\min_i p_i^{(r)} \le D$, we set a shared target length
\begin{equation}
T^{(r)} = \min_i p_i^{(r)} + D .
\end{equation}
Each sample then appends
\begin{equation}
m_i^{(r)} = T^{(r)} - p_i^{(r)}
\end{equation}
mask tokens, where $0 \le m_i^{(r)} \le D$.
Thus, the shortest sample uses the full candidate range, while longer samples append fewer mask tokens and are aligned to the same target length.
All active samples can therefore be processed in a single batched forward pass.

This strategy keeps the candidate range of each sample bounded by $D=32$, while avoiding separate forward passes caused by different PPC commit lengths.
It also preserves the adaptive nature of PPC, since each sample still commits its own reliable prefix according to confidence.
Therefore, batch-level parallelism is maintained without forcing all samples to decode the same number of tokens at each step.

\end{document}